# State Estimation in Visual Inertial Autonomous Helicopter Landing Using Optimisation on Manifold

HOANG Dinh Thinh[1], LE Thi Hong Hieu[2], NGO Dinh Tri[3]

*Abstract*—Autonomous helicopter landing is a challenging task that requires precise information about the aircraft states regarding the helicopter's position, attitude, as well as position of the helipad. To this end, we propose a solution that fuses data from an Inertial Measurement Unit (IMU) and a monocular camera which is capable of detecting helipad's position in the image plane. The algorithm utilises manifold-based nonlinear optimisation over preintegrated IMU measurements and reprojection error in temporally uniformly distributed keyframes, exhibiting good performance in terms of accuracy and being computationally feasible. Our contributions of this paper are the formal address of the landmarks' Jacobian expressions and the adaptation of equality constrained Gauss-Newton method to this specific problem. Numerical simulations on MATLAB/Simulink confirm the validity of given claims.

## I. INTRODUCTION

LANDING is perhaps the most vulnerable phase of autonomous aircraft control and navigation, since any capricious manoeuvre can result in crash and loss of the aircraft. Therefore, the autonomous landing system must be able to function based on highly reliable inputs regarding dynamic states of the aircraft and the position of the landing site. To facilitate the state estimation, Inertial Measurement Unit (IMU) and a monocular camera are cheap and feasible solution that can be combined to yield highly reliable measures of the aircraft attitude, position and location of the landing site. However, major challenges presented include coupling between aircraft - landing site's states, and the inherent uncertainty in both dynamics and sensor measurements.

Several solutions have been proposed to overcome these challenges. They can be divided into two categories, with the first attempted to couple the state estimation and control problem together, while the other left them separated. In particular, the error between the aircraft and the landing site can be calculated directly from camera measures and incorporated as an error state deemed to be regularised by a controller. Representatives of this approach include [1] in which the authors conceived a robust visual target and recognition technique using image moments. From there, the relative yaw angle and position can be inferred.

The other approach typically involves state estimation techniques, and these states are then fed into a feedback controller. In this aspect, either Extended Kalman Filter (EKF) [2], or Particle Filter (PF) are often employed [3]. However, in [2] and [3], the estimation of the landing site is detached from the aircraft's internal states, thus an assumption about the aircraft's attitude that yields little affection on the estimation process had been made.

In this paper, we present a method adapted from [4] that estimates the aircraft's state (attitude, position, velocity) and landmark's position simultaneously using data from IMU and monocular camera. However, different from [4] which adopted the structureless model for visual measurements, we choose to estimate the landmark (i.e. landing site) location directly and incorporate them as an error term in the cost function since this information is valuable for path planning and control system. Optimisation is realised through Levenberg-Marquadt algorithm which is a damped out version of Gauss-Newton, improving robustness when initial condition is far-off from the ground-truth states. We also place additional constrains on the landmark's altitude, which drastically improved the optimisation result.

## II. PROBLEM FORMULATION

### A. States of the system

We denote the state of the system at time $t$ as:
$$X_t = \{R_t, v_t, p_t, p_{l_1 t}, p_{l_2 t}, \ldots, p_{l_N t}\}$$
where $R_t \in SO(3)$ denotes the rotation matrix that transform vector from body frame to world frame, $v_t \in R^3$ is the velocity of the body frame's origin with respect to the world frame, expressed in world's frame, $p_t$ is the position of the body frame origin, expressed in world frame. $N$ landmarks (specific patterns on a helipad) position are denoted $p_{l_i t}$ ($i \in \{1, 2, \ldots, N\}$) expressed in world frame. Assume that the landmarks do not change position with time, we can separate the state vector into two vectors:
$$X_{p_t} = \{R_t, v_t, p_t\}, t \in [0, t_{max})$$
which represents the pose of the aircraft at different camera sample time, and:
$$X_{l_t} = \{p_{l_1}, p_{l_2}, \ldots, p_{l_N}\}$$
which represents the position of the landmarks.

The measurement comes from two sources, one is the IMU which the gyroscope provides angular rates $\widetilde{\omega} \in R^3$, expressed in body frame and acceleration $\tilde{a} \in R^3$, also expressed in body frame. The other is the monocular camera, which, after each image frame is processed and landmarks are recognised, provides coordinates of the landmarks in image plane through the projection equation:
$$p^*_{l_i t} = \pi\left(R_t^T(p_{l_i t} - p_t)\right) = \pi(x_l, y_l, z_l) \quad (1)$$
here, $\pi$ denotes the camera projection equation. To illustrate the basic concept, we adopt the pinhole camera model, such that:
$$x' = f'\frac{x_l}{z_l} \quad (2)$$
and
$$y' = f'\frac{y_l}{z_l} \quad (3)$$
here, x' and y' are the coordinates of the object in image's plane, which correspond to the pixel coordinates of the object in digital image while f' is the focal length of the camera.

### B. Factor graph

The factor graph of Figure 1 illustrates the probabilistic point of view, comprising of a prior state $X_1$, assuming the prior knowledge of the system state, from which the estimation can take place.

[1]Hoang is a graduate student studying at Ho Chi Minh City University Of Technology (e-mail: hdthinh.sdh19@hcmut.edu.vn).
[2]Le is the lecturer of Department of Aerospace Engineering, Ho Chi Minh City University of Technology (e-mail: honghieu.le@hcmut.edu.vn).
[3]Ngo is the lecturer of Department of Control and Automation, Ho Chi Minh City University of Technology (e-mail: ngodinhtri@hcmut.edu.vn).

There are a total number of $n$ pose states $X_p$ in the optimization window and $N$ landmark states, but only $n-1+N$ states needed to be updated, that is all except prior state which correspond to node $X_{p_2}$ to $X_{p_n}$ and $X_{l_1}$ to $X_{l_N}$. The factors, represented as vertices that connect the nodes, comprise of $f_{ij}$, which is the dynamic model that allows next state to be predicted from current state based on IMU data, and $f_k$, which is the measurement model based on camera measurement of landmarks.

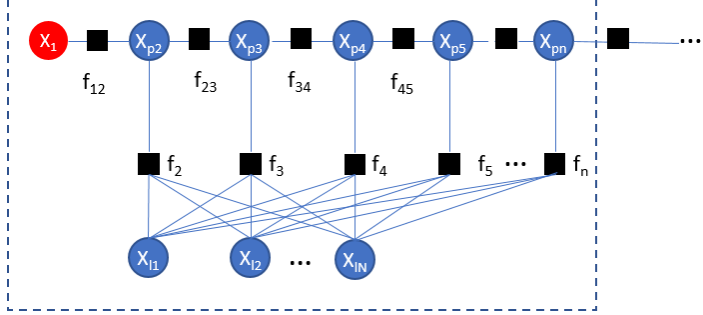

Fig. 1 Factor graph of the problem

In this article, we assume that each state is generated at the same time as new measurements from camera come, while IMU data (usually has higher sampling rate) is accumulated i.e. preintegrated as factor $f_{ij}$ between camera samples.

*C. Residuals and cost function*

The cost function for optimisation is square of residual error, that is:
$$C(X_1, X_2, \ldots, X_t) = e^T W e$$
in which, $e$ is the error residual which include IMU preintegration factors and photometric error, $W$ is a weighting matrix. Several articles such as [5], [6], [4] recommend W to be information matrix, calculated through propagation of the covariance matrix through models that employ uncertainty. However, we propose:
$$W = \begin{bmatrix} 1 & \cdots & 0 \\ \vdots & \ddots & \vdots \\ 0 & \cdots & 1000 \end{bmatrix}$$
that is, an identity matrix but elements associated with photometric error is multiplied by 1000. Such correction is necessary to prevent the optimizer to be biased towards higher order terms.

The goal is to find:
$$X_1, X_2, \ldots, X_t = \underset{X_1, X_2, \ldots, X_t}{\arg\min} C$$
subject to: $z_{l_i} = 0$ where $p_{l_i} = (x_{l_i}, y_{l_i}, z_{l_i})$ is the $i^{th}$ landmark position.

As mentioned earlier, the residuals include IMU residual and photometric error. The IMU residuals for two consecutive frames $i$ and $j = i + 1$ are as follows [4]:
$$r_{\Delta R_{ij}} = Log(R_i^T R_j)$$
$$r_{\Delta v_{ij}} = R_i^T (v_j - v_i - g\Delta t_{ij})$$
$$r_{\Delta p_{ij}} = R_i^T (p_j - p_i - v_i \Delta t_{ij} - \frac{1}{2} g \Delta t_{ij}^2)$$

It is noteworthy that $r_{\Delta v_{ij}}$ and $r_{\Delta p_{ij}}$ do not imply any physical meaning. They are formulated in a way that upcoming measurements can be incorporated independently of the states, alleviating the need for recalculation whenever new measurements arrive [4]:
$$\Delta R_{ij} = R_i^T R_j = \prod_{k=i}^{j-1} Exp((\widetilde{\omega_k})\Delta t)$$
$$\Delta v_{ij} = \sum_{k=i}^{j-1} (\Delta R_{ik}(\widetilde{a_k})\Delta t)$$
$$\Delta p_{ij} = \sum_{k=i}^{j-1} \left( \Delta v_{ik}\Delta t + \frac{1}{2}\Delta R_{ik}(\widetilde{a_k})\Delta t^2 \right)$$

In these expressions, the logarithm map (Log) is defined the same as the logarithm map of the SO(3) group, except with a slight abuse of notation:
$$Log: SO(3) \rightarrow R^3$$
$$R \rightarrow \frac{\phi(R - R^T)}{2\sin(\phi)}$$
in which:
$$\phi = \arccos\left(\frac{trace(R) - 1}{2}\right)$$
while:
$$Exp: R^3 \rightarrow SO(3)$$
$$R = I + \frac{\sin(\phi)}{\phi}\phi^\wedge + \frac{1-\cos(\phi)}{\phi^2}\phi^{\wedge 2}$$
The "hat map" is defined as:
$$\wedge: R^3 \rightarrow so(3)$$
$$\begin{bmatrix} \omega_1 \\ \omega_2 \\ \omega_3 \end{bmatrix} \rightarrow \begin{bmatrix} 0 & -\omega_3 & \omega_2 \\ \omega_3 & 0 & -\omega_1 \\ -\omega_2 & \omega_1 & 0 \end{bmatrix}$$
The photometric residual is defined as:
$$e_{photo_k} = \begin{bmatrix} \hat{x}' - x' \\ \hat{y}' - y' \end{bmatrix}$$
in which, $\hat{x}'$ and $\hat{y}'$ are predicted position of the landmark $k$ in the image plane, calculated from equation (2) and (3), while x' and y' are the measured landmark position given by the feature detector from camera image.

III. OPTIMISATION, DERIVATION OF THE JACOBIAN OF PHOTOMETRIC ERROR AND LANDMARK ALTITUDE CONSTRAIN

*A. Optimisation*

The optimisation employs Levenberg-Marquadt scheme, which is a damped out version of Gauss-Newton optimisation for least-square nonlinear problem. The method approximates the cost function around the neighborhood of current estimate by a second-order Taylor series. Then, an optimal update to minimize the approximated cost function is made, and the process is repeated until convergence.

Hessian is first calculated from Jacobian matrix $J$:
$$H(X) = J(X)^T W J(X) + \alpha I \quad (4)$$
and
$$b(X) = J(X)^T W e(X) \quad (5)$$
the update is calculate as:
$$\delta = H^{-1} b$$
$$X_{new} = X \boxplus \delta \quad (6)$$

The $\boxplus$ operator is slightly different from traditional addition operator in vector space. It is defined traditional addition operator for all variables, except for the rotation matrix and the velocity of the body frame which:
$$X_t^*: R^* = R Exp(\delta_R)$$
$$p^* = p + R\delta_p$$

In this aspect, we have "lifted" the optimisation on manifold $(SO(3)^p \times R^q)$ to the vector space $R^{q+3p}$. Since traditional least-square optimisation methods only work in vector space, lifting made it possible to employ such methods for manifolds that maintain the original structure and keep the manipulation simple.

Expressions of the Jacobian of IMU preintegrated residual are given in [4]. However, [4] utilises the Schur-complement trick for structureless vision, while the landmarks' location is actually important in autonomous helicopter landing. Therefore, we propose a derivation of such fomulas in the next section.

### B. Derivation of the Jacobian of Photometric error

For derivation of $\frac{d}{d\delta_r} p^*_{l\,i_t}(X \boxplus \delta)$ in which:
$$\left(RExp(\delta_r)\right)^T (p_l - p) = Exp(-\delta_r)\, R^T (p_l - p)$$
If $\delta_r$ is small, we can approximate:
$$Exp(-\delta_r) \approx I - \delta_r{}^\wedge$$
Thus:
$$= \frac{d}{d\delta_r}(I - \delta r^\wedge) R^T (p_l - p) = \frac{d}{d\delta_r}(-\delta r^\wedge (R^T(p_l - p)))$$
Since the hat map is equivalent to the cross-product:
$$a^\wedge b = a \times b = -b \times a = -b^\wedge a$$
We can rewrite the expression as:
$$= \frac{d}{d\delta_r}(R^T(p_l - p))^\wedge \delta r$$
Hence:
$$\frac{d}{d\delta_r} p^*_{l\,i_t}(X \boxplus \delta) = (R^T(p_l - p))^\wedge$$

The expression of (1) does not involve velocity, thus changes to velocity of the states during optimization does not affect the photometric residual error.
$$\frac{d}{d\delta_v} p^*_{l\,i_t}(X \boxplus \delta) = 0_{3\times 3}$$

For updates to the landmark position, as formerly stated, the new landmark position is:
$$p^*_{l_i} = p_{l_i} + \delta p_{l_i}$$
Thus:
$$\frac{d}{d\delta p_{l_i}} p^*_{l\,i_t}(X \boxplus \delta) = \frac{d}{d\delta p_{l_i}}\left(R^T(p_{l_i} + \delta p_{l_i} - p)\right) = R^T$$
For updates to aircraft's position, the formula is:
$$\frac{d}{d\delta p} p^*_{l\,i_t}(X \boxplus \delta) = \frac{d}{d\delta p}\left(R^T(p_{l_i} - p - R\delta p)\right) = -I_{3\times 3}$$

We then use the chain rule to derive the formua for $e_{photo_k}$. Note that:
$$\frac{d}{d\delta} p^*_{l\,i_t} = \begin{bmatrix} \frac{dx_l}{d\delta_1} & \frac{dx_l}{d\delta_2} & \cdots \\ \frac{dy_l}{d\delta_1} & \frac{dy_l}{d\delta_2} & \cdots \\ \frac{dz_l}{d\delta_1} & \frac{dz_l}{d\delta_2} & \cdots \end{bmatrix}$$
in which, $\delta = [\delta_1, \delta_2, \delta_3, \ldots]^T$. Then:
$$\frac{d}{d\delta} e_{photo_k} = f' \begin{bmatrix} \frac{\frac{dx_l}{d\delta_1} z_l - \frac{dz_l}{d\delta_1} x_l}{z_l^2} & \frac{\frac{dx_l}{d\delta_2} z_l - \frac{dz_l}{d\delta_2} x_l}{z_l^2} & \cdots \\ \frac{\frac{dy_l}{d\delta_1} z_l - \frac{dy_l}{d\delta_1} y_l}{z_l^2} & \frac{\frac{dy_l}{d\delta_2} z_l - \frac{dy_l}{d\delta_2} y_l}{z_l^2} & \cdots \end{bmatrix}$$

### C. Constraining the landmark altitude

We place additional constrains regarding the landmark altitude (which is assumed to be known as 0). These constrains allow faster convergence, an require less landmarks in order to yield adequate accuracy. Denote $J_h$ to be the Jacobian of $z_{l_i}$, which is trivially computed as $[0\ 0\ 0\ \ldots 0\ 1\ 0\ \ldots]$ where the position of "1" corresponds to the position of the height of landmark position in $p_{l_t} \in X_t$. Equality constrains mean the new system is:
$$\begin{bmatrix} H & J_h^T \\ J_h & 0 \end{bmatrix} \begin{bmatrix} \delta \\ \lambda \end{bmatrix} = \begin{bmatrix} -J^T We \\ 0 \end{bmatrix}$$

Then, $\delta$ can be extracted and perform update in the same way as (6).

## IV. NUMERICAL SIMULATION

### A. Determining number of window length and landmarks

The sufficient condition for $H(X)$ to be invertible is that $J$ achieves full-column rank. However, it is usually impossible to conceive a flight path that always yield such condition, thus we only deal with necessary condition rather than sufficient condition for H to be invertible.

$J$, for $N$ landmarks and window length of $n$, has:
$$(n - 1) \times 9 + n \times N \times 2 \text{ rows}$$
$$n \times 9 + N \times 3 \text{ columns}$$

The necessary condition for $J$ to achieve full rank is when the number of rows is larger than the number of columns:
$$n(n - 1) \times 9 + n \times N \times 2 - n \times 9 + N \times 3 > 0$$
yielding:
$$N > \frac{9}{2n - 3}$$

Consequently, for a window size of 6, we only need one landmark. However, in practice, we may need much higher number of landmarks to reduce interference from outliers and uncertainties of measurements.

### B. Numerical experiment

We conduct a numerical simulation consisting of 3 landmarks with a window size of 7. Initial condition is depicted in Table I. A quadrotor originally at X=0m, Y=0m and Z=4m begins to fly a trajectory depicted in Figure 2, observing 3 landmarks (of the helipad, which is 3 AprilTags [7]). Quadrotor nonlinear geometric control [8] is employed for trajectory tracking.

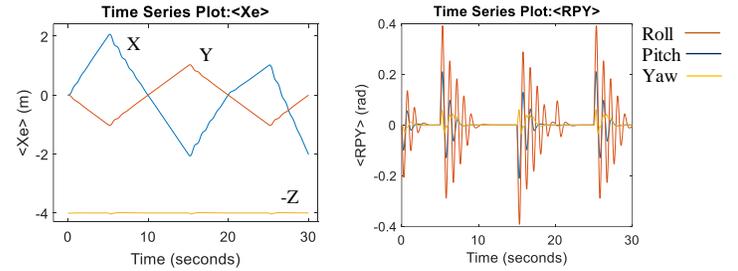

(a) Position of aircraft  (b) Roll, pitch, yaw angle
Fig. 2 Quadrotor trajectory tracking performance

TABLE I
EXPERIMENT SETUP FOR NUMERICAL SIMULATION

| Quantity | Value | Unit |
| --- | --- | --- |
| IMU measures between camera samples | 20 | - |
| Levenberg-Marquadt damping constant $\alpha$ | 0.1 | - |
| IMU sampling time | 0.02 | s |
| Camera sampling time (Includes signal processing delay) | 0.4 | s |
| Optimisation iterations | 50 | |
| Processing time* | 3.73 | s |
| Initial attitude (for optimisation) | I | |
| Initial position (for optimisation) | 0, 0, -4 | m |
| Initial velocity (for optimisation) | 0, 0, 0 | m/s |

* Processing time is measured on an Intel Core i5 2500K CPU, 8GB RAM, no GPU acceleration.

In the simulation, noise power of IMU is set to 0.0001 for all components. After 50 iterations, the total cost function is minimised to chatter around 0.2 (Figure 3).

Table II, III and IV compare the final optimisation result with respect to ground-truth states.

## V. Discussion

Data accumulated from 140 IMU samples and 7 camera samples are enough to give adequate accuracy regarding system states and landmark positions. The estimation error was about less than 30cm for aircraft's position and 20cm for landmark positions. It is noteworthy to point out that the last camera sample in aircraft state usually accumulate higher error, due to the fact that it is less constrained, i.e. has only one factor connecting to it (Figure 1) than others. However, outliers and increased noise in real environment can contribute to increasing error when tested in real life scenarios. Nevertheless, it is believed that the algorithm will perform well, as many SLAM systems do [4] [9].

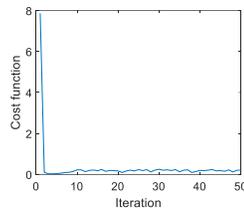

Fig. 3 Quadrotor trajectory tracking performance

TABLE II

DIFFERENCE IN POSITION WITH RESPECT TO GROUND TRUTH STATE

| Camera sample | X (m) | Y (m) | Z(m) |
|---|---|---|---|
| 1 | 0.0000 | 0.0000 | 0.0000 |
| 2 | 0.0012 | 0.0033 | 0.0010 |
| 3 | -0.0336 | 0.0112 | 0.0070 |
| 4 | -0.1297 | 0.0268 | 0.0163 |
| 5 | -0.2043 | 0.0709 | 0.0240 |
| 6 | -0.2881 | 0.1504 | 0.0317 |
| 7 | -0.4952 | 0.2808 | 0.0420 |

TABLE III

DIFFERENCE IN LANDMARK ESTIMATE WITH RESPECT TO GROUND TRUTH STATE

| Landmark | X (m) | Y (m) | Z(m) |
|---|---|---|---|
| 1 | 0.0558 | -0.0132 | 0.0000 |
| 2 | 0.0446 | -0.0279 | 0.0000 |
| 3 | 0.0355 | -0.0035 | 0.0000 |

Trials with higher IMU preintegrated measures in each factor, larger number of landmarks all led to better performance. It is advised that the Levenberg-Marquadt damping factor is kept low, about 0.1 to 0.2 to yield adequate optimisation result. Although not employed in this paper, marginalisation of older states from the optimisation window may improve the result and keep the computational demand constant [10].

## VI. Conclusion

In this paper, we have presented a method to simultaneously estimate the aircraft state and position of detected landmarks from the camera. Insights from these findings will be useful, not only to construct a fully autonomous landing system for helicopters and drones, but to also design a Simultaneous Localisation and Mapping (SLAM) algorithm. Further testing with in life scenarios is warranted in order to study the robustness of the algorithm.